\documentclass{article}
\usepackage{ijcai24}

\usepackage{times}
\usepackage{soul}
\usepackage{url}
\usepackage[hidelinks]{hyperref}
\usepackage[utf8]{inputenc}
\usepackage[small]{caption}
\usepackage{graphicx}
\usepackage{amsmath}
\usepackage{amsthm}
\usepackage{booktabs}
\usepackage{algorithm}
\usepackage{algorithmic}
\usepackage[switch]{lineno}
\usepackage{xcolor}

\usepackage{multirow}
\usepackage{enumerate}
\usepackage{dsfont}
\usepackage{arydshln}
\usepackage{pifont}
\usepackage{caption}
\usepackage{xspace}
\usepackage{pdfpages}

\usepackage{soul}

\usepackage{bm}

\title{UniM-OV3D: Uni-Modality Open-Vocabulary 3D Scene Understanding with Fine-Grained Feature Representation}

\author{
Qingdong He$^1$\footnote{Equal contribution.}\and
Jinlong Peng$^{1*}$\and
Zhengkai Jiang$^1$\and
Kai Wu$^1$\and
Xiaozhong Ji$^1$\and \\
Jiangning Zhang$^1$\footnote{Corresponding author.}\addtocounter{footnote}{-1}\stepcounter{mpfootnote}\and 
Yabiao Wang$^1$\footnotemark\and 
Chengjie Wang$^1$\and 
Mingang Chen$^2$\and 
Yunsheng Wu$^1$
\affiliations
$^1$YouTu Lab, Tencent \hspace{0.4cm}
$^2$Shanghai Development Center of Computer Software Technology
\emails
\{yingcaihe, jeromepeng, zhengkjiang, lloydwu, xiaozhongji, vtzhang, caseywang, jasoncjwang,simonwu\}@tencent.com,
cmg@sscenter.sh.cn
}

\begin{document}

\maketitle

\begin{abstract}
    3D open-vocabulary scene understanding aims to recognize arbitrary novel categories beyond the base label space. However, existing works not only fail to fully utilize all the available modal information in the 3D domain but also lack sufficient granularity in representing the features of each modality. In this paper, we propose a unified multimodal 3D open-vocabulary scene understanding network, namely UniM-OV3D, which aligns point clouds with image, language and depth. To better integrate global and local features of the point clouds, we design a hierarchical point cloud feature extraction module that learns comprehensive fine-grained feature representations. Further, to facilitate the learning of coarse-to-fine point-semantic representations from captions, we propose the utilization of hierarchical 3D caption pairs, capitalizing on geometric constraints across various viewpoints of 3D scenes. Extensive experimental results demonstrate the effectiveness and superiority of our method in open-vocabulary semantic and instance segmentation, which achieves state-of-the-art performance on both indoor and outdoor benchmarks such as ScanNet, ScanNet200, S3IDS and nuScenes. Code is available at \textit{\textcolor{magenta}{\url{https://github.com/hithqd/UniM-OV3D}}}.
    
\end{abstract}

\section{Introduction}
Accurate and resilient comprehension of 3D scenes plays a vital role in various practical applications, including autonomous driving~\cite{mao20223d}, virtual reality~\cite{park2008multiple} and robot navigation~\cite{shafiullah2022clip}. Despite significant progress in recognizing closed-set categories on standard datasets~\cite{graham20183d,vu2022softgroup,misra2021end}, existing models frequently struggle to recognize novel categories that are not present in the training data label space. The limited scalability of existing models in open-set scenarios hinders their practical applicability in the real world and motivates us to explore the open-vocabulary 3D scene understanding capability. 

\begin{figure}[t]
    \centering
\includegraphics[width=0.95\linewidth]{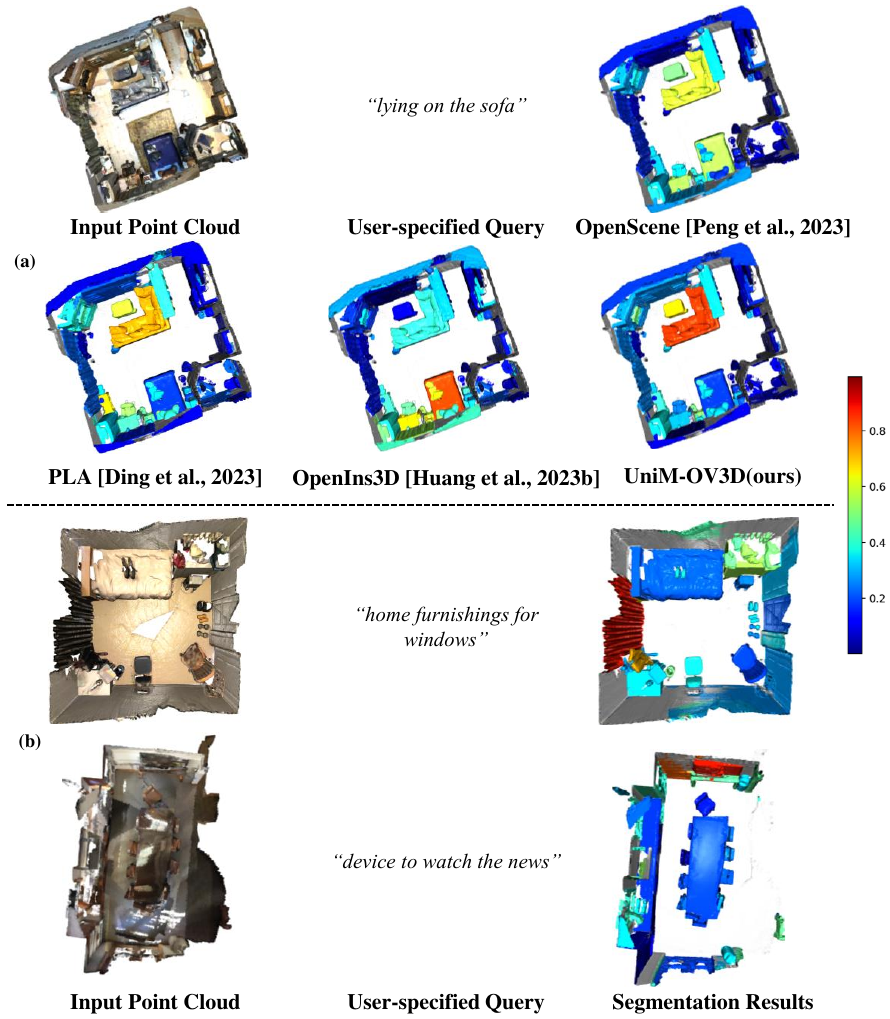}
    \caption{Open-vocabulary 3D scene understanding. Different colors represent the confidence of matching the user-specified query. (a) Comparison of different methods using the same query, (b) Results of our method in complex reasoning or content that requires extensive world knowledge.}
    \label{fig:1}
\end{figure}

Benefit from the 2D foundation model, such as CLIP~\cite{radford2021learning} and DINO~\cite{caron2021emerging}, previous methods~\cite{peng2023openscene,chen2023clip2scene,liu2023weakly,liu2023openshape} attempt to distill knowledge from 2D pixels to 3D features, aligning point clouds with 2D images. Another line of works~\cite{takmaz2023openmask3d,huang2023openins3d} focuses on points-only mechanism. The existing methods inadequately investigate data from multiple modalities, such as point clouds, images, linguistic input, and depth information, which are crucial for effective open-vocabulary 3D scene understanding.
Despite their effectiveness, the absence of alignment and learning for other modalities makes them miss a lot of semantic information and hinders their ability to effectively handle fine-grained point cloud object instances, as shown in Figure~\ref{fig:1} (a). 

Considering the properties of 3D scene, we should not overlook the inherent characteristics of 3D data itself. Firstly, depth information serves as a crucial modality for depth-invariant feature aggregation, which is often neglected.
Early works~\cite{zhang2022pointclip,zhu2023pointclip,huang2023clip2point} in 3D recognition have attempted to transfer depth modality to 3D understanding. 
However, mere projection or rendering, and merely aligning with a single modality such as image or text, cannot fully exploit all the characteristics of depth information, nor can it perform well in the task of open-vocabulary 3D scene understanding. 
The second aspect to consider is the exploration of point cloud information itself. Aiming at the point clouds caption learning, previous approaches either solely rely on simplistic templates derived from CLIP~\cite{peng2023openscene,liu2023weakly,zhang2023clip} or utilize images as a bridge to generate corresponding textual descriptions from 2D images~\cite{ding2023pla,yang2023regionplc}. 
However, point clouds captioning can make alignment easier and more comprehensive, particularly when multiple modalities are incorporated during the alignment training.
And the point clouds encoder almost comes from mature 3D extractors~\cite{choy20194d,graham20183d}, which often keep frozen during training.
The frozen backbone and bridging alignment impede the generalization in complex 3D open vocabulary scenarios.

To this end, in order to fully leverage the synergistic advantages of various modalities, we propose a comprehensive multimodal alignment approach that co-embeds 3D points, image pixels, depth, and text strings into a unified latent space for open-vocabulary 3D scene understanding, namely UniM-OV3D. We thoroughly explore and utilize the characteristics of point clouds from two perspectives. First, to extract fine-grained geometric features of different levels from the original irregular 3D point clouds, we design a hierarchical point cloud feature extraction module. 
Second, in terms of generating point-caption pairs, we make the first attempt to generate corresponding text directly from point clouds instead of using images as a bridge. And we build hierarchical point-semantic caption pairs, including global-, eye- and
sector-view captions, which can offer fine-grained language supervisions.

Furthermore, we introduce depth information into the entire network and the modified depth encoder inspired by CLIP2Point~\cite{huang2023clip2point} keeps learning during training . As for image modality, we employ the pre-trained model PointBIND~\cite{guo2023point} to extract image feature. Given the multi-modalities and their dense representations, we finally conduct multimodal contrastive learning for robust 3D understanding. In this way, we are capable of optimizing their reciprocal benefits by adeptly amalgamating multiple modalities. 

Extensive experiments on ScanNet~\cite{dai2017scannet}, ScanNet200~\cite{rozenberszki2022language}, S3DIS~\cite{armeni20163d} and nuScenes~\cite{caesar2020nuscenes}  show the effectiveness of our method on 3D open-vocabulary tasks, surpassing previous state-of-the-art methods by 3.2$\%$-7.8$\%$ hIoU on semantic segmentation and 3.8$\%$-10.8$\%$ hAP$_{50}$ on instance segmentation. More interestingly, UniM-OV3D demonstrates a robust ability to understand complex language queries, even those that involve intricate reasoning or necessitate extensive world knowledge, as illustrated in Figure~\ref{fig:1} (b).

Our main contributions are summarized as follows:
\begin{itemize}
    \item Within a joint embedding space, UniM-OV3D firstly aligns 3D point clouds with multi-modalities, including 2D images, language, depth, for robust 3D open-vocabulary scene understanding.
    \item In order to acquire more comprehensive and intricate fine-grained geometric features from point clouds, we propose a hierarchical point cloud extractor that effectively captures both local and global features.
    \item We innovatively build hierarchical point-semantic caption pairs that offer coarse-to-fine supervision signals, facilitating learning adequate point-caption representations from various 3D viewpoints directly. 
    \item UniM-OV3D outperforms previous state-of-the-art methods on 3D open-vocabulary semantic and instance segmentation tasks by a large margin, covering both indoor and outdoor scenarios.
\end{itemize}

\begin{figure*}[ht]
\centering
\includegraphics[width=0.95\linewidth]{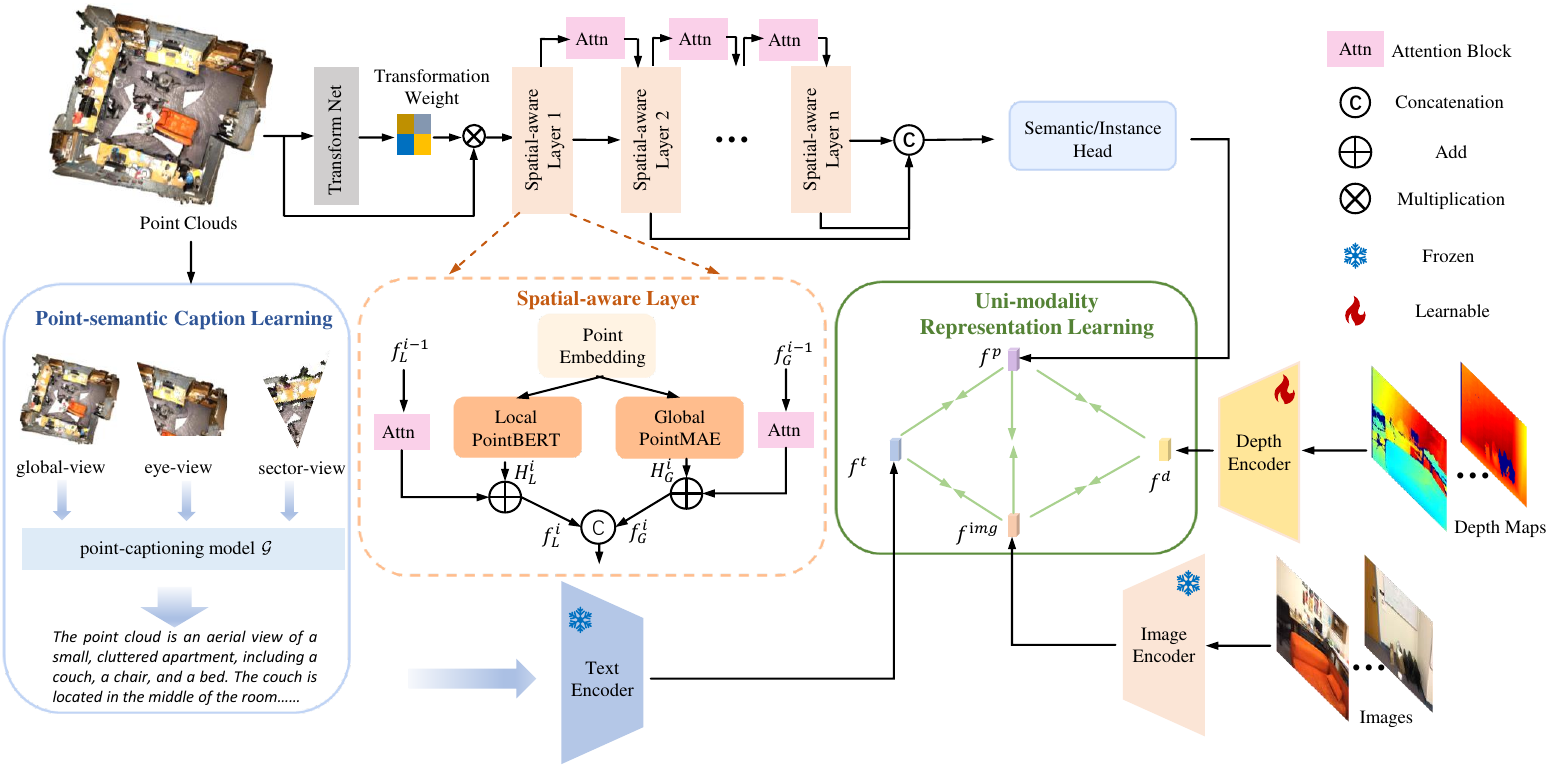}
\caption{\textbf{Architecture of our proposed UniM-OV3D.} The input point clouds are processed by a hierarchical point cloud extraction module to fuse the local and global features. To fulfill coarse-to-fine text supervision signal, the point-semantic caption learning is designed to acquire representations from various 3D viewpoints. The overall framework takes point clouds, 2D image, text and depth map as input to establish a unified multimodal contrastive learning for open-vocabulary 3D scene understanding.}
\label{fig:2}
\end{figure*}
\section{Related Work}
\subsection{3D Recognition}
To process point clouds directly, PointNet~\cite{qi2017pointnet} and PointNet++~\cite{qi2017pointnet++} are the two pioneering studies in this field that focus on developing parallel MLPs for feature extraction from unstructured data which have significantly enhanced accuracy~\cite{huang2023learning,huang20223qnet}. Further, while the VLM~\cite{radford2021learning,caron2021emerging} has achieved remarkable results in zero-shot or few-shot learning of 2D images by utilizing extensive image data, the availability of such internet-scale data for 3D point clouds is limited. Consequently, extending the open-vocabulary mechanism to 3D perception is a nontrivial task due to the difficulty in gathering sufficient data for point-language contrastive training, similar to the approach employed by CLIP~\cite{radford2021learning} in 2D perception. Therefore, PointCLIPs~\cite{zhang2022pointclip,zhu2023pointclip} take the first step to convert point clouds into CLIP-recognizable images and align the point cloud feature  with the language feature by projecting 3D point cloud into 2D space. 
Similarly, CLIP2Point~\cite{huang2023clip2point} attempts to align the projected depth map with the images by designing a trainable depth encoder and cross-modality learning losses. Different from point projecting, a series of works~\cite{xue2023ulip,xue2023ulip2,zeng2023clip2,liu2023openshape} try to collect multi-modalities triplets to train the 3D backbone. To enable zero-shot capability and improve standard 3D recognition, they align the 3D feature with the CLIP-aligned visual and textual features. This alignment facilitates the integration of 3D features with CLIP, enhancing both the zero-shot capability and the standard 3D recognition capability.
However, existing methods cannot integrate the four modal information types, namely point cloud, image, depth, and text, which is the focus of this paper. We propose a unified multimodal network that incorporates fine-grained feature representation from these four modalities, effectively leveraging their strengths.

\subsection{Open-Vocabulary 3D Scene Understanding}
Facing the challenge of the lack of diverse 3D open-vocabulary datasets for training generalizable models, one solution is to distill knowledge between pre-trained 2D open-vocabulary models and 3D models~\cite{wu2023towards}. OpenScene~\cite{peng2023openscene} and CLIP-FO3D~\cite{zhang2023clip} propose to distill the knowledge of 2D CLIP into 3D feature without any extra annotation. OpenMask3D~\cite{takmaz2023openmask3d} first generates class-agnostic instance mask proposals with the 3D point cloud, which is used to choose the 2D views and masks. Meanwhile, 3D-OVS~\cite{liu2023weakly} distills the open-vocabulary multimodal knowledge and object reasoning capability of CLIP into a neural radiance field. OpenIns3D~\cite{huang2023openins3d} employs a “Mask-Snap-Lookup” scheme to learn class-agnostic mask proposals in 3D point clouds. Another line focuses on point-language contrastive training. To obtain 3D backbone-language alignment, PLA-family~\cite{ding2023pla,yang2023regionplc,ding2023lowis3d} uses 2D images as a bridge to generate descriptions of point cloud data. Although carefully designed, describing point clouds through images may not fully capture the details and point-semantic information due to differences between point clouds and images. Therefore, methods that describe point cloud data more directly and accurately need to be further explored. So we establish a point-semantic caption learning mechanism, in which captions can be obtained directly from point clouds with more accurate coarse-to-fine languages supervisions.

\section{Method}

\subsection{Preliminary}\label{sec:3.1}
The objective of open-vocabulary 3D scene understanding is to localize and recognize unseen categories without the need for corresponding manual annotations. To put it formally, annotations on semantic and instance levels, denoted as \begin{math} \mathcal{Y} \end{math} $=\{y^B,y^N\}$, are split into base categories \begin{math} \mathcal{C} \end{math}$^B$ and novel categories \begin{math} \mathcal{C} \end{math}$^N$. During the training phase, the 3D model has access to all point clouds \begin{math} \mathcal{P} \end{math} $=\{\mathbf{p}\}$, but only to annotations for base classes \begin{math} \mathcal{C} \end{math}$^B$. It remains unaware of both annotations $y^N$ and category names related to novel classes \begin{math} \mathcal{C} \end{math}$^N$. However, during the inference process, the 3D model is required to localize objects and classify points belonging to both basic and novel categories \begin{math} \mathcal{C} \end{math}$^B$$\cup$ \begin{math} \mathcal{C} \end{math}$^N$. This inference step is crucial for achieving comprehensive scene understanding and enabling the model to generalize to unseen classes. During inference, the model can predict any desired category by computing the similarity between point-wise features and the embedding of queried categories, enabling open-world inference.

\subsection{Hierarchical Feature Extractor with Local and Global Fusion}\label{sec:3.2}
Taking the sparse point clouds as input, we propose a trainable hierarchical point cloud extractor to capture fine-grained local and global features, instead of just utilizing the frozen 3D extractors. As illustrated in Figure~\ref{fig:2}, the input is channeled into a transformer network, which employs attention-based layers to regress a $4\times4$ transformation matrix. This matrix comprises the elements that represent the learned affine transformation values, which are used for aligning point clouds. Following alignment, the points are introduced into multiple stacked spatial-aware layers, which serve to produce a permutation-invariant embedding of these points. Within this structure, the residual attention module functions as a linking bridge between two adjacent layers, facilitating the transfer of information. After the information has been processed through the attention-based layers, the outputs from all these $N$-dimensional layers are concatenated. Finally, the segmentation head can be added to output the global information aggregation for the point clouds, providing a comprehensive summary of the data.

\paragraph{Spatial-aware Layers.} The architecture of the spatial-aware layers is shown in Figure~\ref{fig:2}, where the current $i$ layer acts as an intermediate layer and the features are not only transmitted by the mainstream information flow but the residual attention linking from $i-1$ layer. Consider a $R^D$ dimensional embedding point clouds set with $n$ points in the $i$ layer $\mathbf{p}=\{p_{i,1},p_{i,2},...,p_{i,n}\}$. In each attention-based flow, the local PointBERT~\cite{yu2022point} layer and the global PointMAE~\cite{pang2022masked} layer process these points in parallel. In the branch of the local PointBERT layer, a transformation is employed on the points to get the feature $H_L^i$:
\begin{equation}
    H_L^i=\{h_{i,1}^M,h_{i,2}^M,...,h_{i,g}^M\} = B(p_{i,1},p_{i,2},...,p_{i,n})
\end{equation}
where $B:R^D\rightarrow R^M$ and $g$ is the local patch. After that, the attention map from $i-1$ layer by the residual linking is added to the output $H_L^{i}$ in a point-wise manner which can be written as:
\begin{equation}
    f_L^i = H_L^i +  \omega_1 \otimes H_L^{i-1}
\end{equation}
where $\otimes$ denotes element-wise multiplication. $\omega_1$ functions as a feature filter, dampening the impact of noisy features while amplifying the beneficial ones. Similarly, global PointMAE layer branch applies global transformation $E$ on the points $\mathbf{p}$ and the output from decoder denotes $H_G^i$. And the final output after the linking of attention block is:
\begin{equation}
    f_G^i = H_G^i +  \omega_2 \otimes H_G^{i-1}
\end{equation}
where $\omega_2$ is similar to $\omega_1$. The outputs $f_L^i$ and $f_G^i$ from the two branches are of the same dimension and are transferred to the attention block of the next layer. Ultimately, they are concatenated together, and these combined embedding are used as the input for the subsequent layer. 

The attention block consists of a series of residual units~\cite{he2016identity}, inserted with interpolation to form a symmetrical top-down architecture. Two consecutive $1\times1$ convolution layers are connected at the end to balance the dimensions. More details are presented in the supplementary.


\begin{figure}[t]
    \centering
\includegraphics[width=\linewidth]{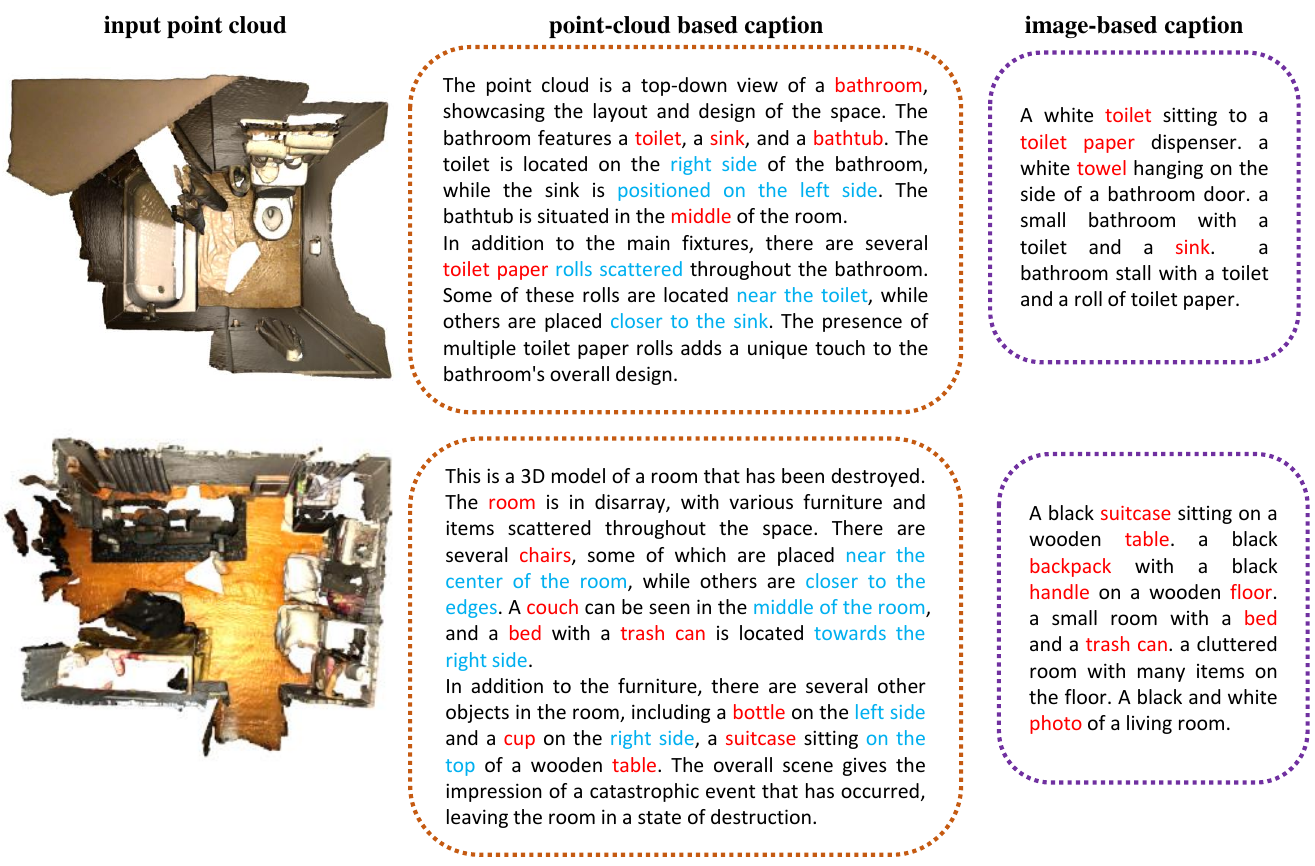}
 \vspace{-0.5cm}
\caption{Point-caption pairs comparison. The middle column represents the captions generated by the point-captioning model, PointLLM~\protect\cite{xu2023pointllm}, and the third column represents the corresponding image captions.}
\label{fig:4}
\vspace{-0.4cm}
\end{figure}

\subsection{Point-semantic Caption Learning}\label{sec:3.3}

Recent success of open-vocabulary works in 2D~\cite{bangalath2022bridging,wu2023diffumask} and 3D~\cite{ding2023pla,yang2023regionplc} vision has shown the effectiveness of introducing language supervision to guide the model to access abundant semantic concepts with a large vocabulary size. However, utilizing images as a bridge to associate the points with language is the core idea in these 3D works. And the final textual descriptions in the so-called point-caption pairs are all sourced from image descriptions and do not possess semantic information specific to the point cloud. To address this problem, we propose to provide direct 
language supervision and a coarse-to-fine point-semantic caption generation mechanism is designed.

\paragraph{Global-view Point-semantic Caption.} Due to the similarity of point-captioning and image captioning~\cite{mokady2021clipcap,wang2022ofa} as fundamental tasks, various base models~\cite{xu2023pointllm,guo2023point} trained on a large number of 3D point samples are already available for perceiving 3D scenes. Given the $i^{th}$ 3D point cloud scene $\mathbf{p}_{i}$, the simplest and most rudimentary manner is to link language supervision to all points. Specifically, the corresponding global language description ${\mathbf{t}}_{i}^{global}$ can be generated by the pre-trained point-captioning model \begin{math} \mathcal{G} \end{math} as follows,
\begin{equation}
    {\mathbf{t}}_{i}^{global} = \mathcal{G}(\mathbf{p}_{i}).
\end{equation}
Surprisingly, with the semantic understanding on the overall points, the entities covered in the generated captions have already encompassed the entire 3D semantic label space. As illustrated in Figure~\ref{fig:4}, the middle column represents the captions generated by the point-captioning model, and the third column represents the corresponding image captions. It is observable that the global point-captions not only offer a more precise and holistic depiction of the scene, but they also represent the orientation information of objects more accurately within the scene and the interrelationships among them. Despite the simplicity of {\tt global-view} caption, we find that it can significantly improve the whole open vocabulary capabilities of understanding the scene and locating objects.

\paragraph{Hierarchical Point-semantic Caption.} Diving into the exploration of more fine-grained point-semantic caption learning, {\tt global-view} point cloud captions are still too coarse to capture fine-grained details between objects in point clouds, making them suboptimal for 3D scene understanding tasks. Hierarchical point captions can further refine the semantic information of point clouds without ignoring all object features in the scene. Therefore, we propose two other association patterns on point sets at different 3D viewpoints.

For each 3D point cloud scene $\mathbf{p}_{i}$, we initiate from the x-axis and segment the entire point cloud space into sector-shaped regions with an angular magnitude of $\theta$ degree, where the intersecting angle between adjacent sectors is $\phi$ degree. The segmented 3D scene $\mathbf{p}_{i}^s$ can be denoted as:
\begin{equation}
    \mathbf{p}_{i}^s = \{p_{ij}^{\theta\_\phi} \mid 0 \leq \phi<\theta, \theta> 0, 0< j < \lceil {\frac{360}{\theta}} \rceil \}.
\end{equation}
For the partitioned sector-shaped regions $p_{ij}^{\theta\_\phi}$, we categorize them into two distinct types of viewpoints based on the magnitude of the angle, named as {\tt sector-view} and {\tt eye-view}. The corresponding angle size of the two views is $\theta_1$ and $\theta_2$, where $0 < {\theta_1} \leq {90}$ and $90< \theta_2 \leq 180$. This implies that the {\tt sector-view} focuses on the minutiae of local point cloud information, while the {\tt eye-view} concentrates on a broader point cloud area, yet the extent of this area is within an obtuse angle range, akin to human eyes. 

\paragraph{Contrastive Point-semantic Caption Training.} With the obtained {\tt global-}, {\tt eye-} and {\tt sector-view} point-semantic captions from different views, we can conduct point-language feature contrastive learning to guide the 3D network to learn from vocabulary-rich language supervisions. 

Specifically, we extract the text embeddings $\mathbf{f}_i^t$ based on the pre-aligned and fixed text encoder in PointBIND~\cite{guo2023point}. And the 3D feature $\mathbf{f}_i^p$ can be acquired by the trainable 3D point cloud encoder. We formulate the point-to-text alignment using the contrastive loss as:
\begin{equation}\label{eq:capt}
     {\mathcal{L}}_{capt}^m =-\frac{1}{N}\sum_{i=1}^{N}log{\frac{\exp(\mathbf{f}_i^p \cdot \mathbf{f}_i^t /\tau )}{\sum _{j=0}^{N}\exp(\mathbf{f}_i^p \cdot \mathbf{f}_j^t/\tau) } } 
\end{equation}
where $m \in \{global, eye, sector\}$, $N$ is the total number of the point-semantic caption pairs and $ \tau$ is a learnable temperature parameter. With Equation~\ref{eq:capt}, we can easily compute different caption losses on {\tt global-view} ${\mathcal{L}}_{capt}^{global}$, {\tt eye-view} ${\mathcal{L}}_{capt}^{eye}$ and {\tt sector-view} ${\mathcal{L}}_{capt}^{sector}$. The final caption loss is a weighted combination between different views as follows:
\begin{equation}\label{eq:capt_total}
     \mathcal{L}_{capt}^{total} = \alpha{\mathcal{L}}_{capt}^{global}+ \beta{\mathcal{L}}_{capt}^{eye} + \gamma{\mathcal{L}}_{capt}^{sector}
\end{equation}
where $\alpha$, $\beta$ and $\gamma$ are used to balance the relative importance of different parts and are set to 1, 0.8, 0.8 by default. The effect of different views and the comparison of different combinations are shown in ablation studies.

\subsection{Aligning Multimodal Representations}\label{sec:3.4}
With the construction of the four modalities, the subsequent objective of UniM-OV3D is to conduct dense alignment between the 3D points with their corresponding image, depth and text.

For the paired image-text ($\textbf{I}$, $\mathbf{t}$) data, we utilize their corresponding encoders from PointBIND~\cite{guo2023point} for feature extraction, which are kept frozen during training. For each depth map \begin{math} \mathcal{D} \end{math}, we set the gated aggregator weights to 0 in CLIP2Point~\cite{huang2023clip2point} to extract the depth feature, formulated as
\begin{align}\label{eq:imgtextgen}
     \mathbf{f}^t, \mathbf{f}^{img} & = PointBIND(\mathbf{t},\textbf{I}) \nonumber \\
     & \mathbf{f}^d = E_d(\mathcal{D})
\end{align}
where $\mathbf{f}^t$, $\mathbf{f}^{img}$, $\mathbf{f}^d$ denote the text, image and depth embeddings. And the depth encoder $E_d$ keeps learnable during training. For the 3D point clouds, we adopt the designed hierarchical point cloud feature extractor module to extract 3D features for the point cloud. In order to transform the encoded 3D feature into multi-modal embedding space, we append a projection network after the 3D encoder $E_p$. So the final 3D features $\mathbf{f}^p$ can be formulate as :
\begin{equation}\label{eq:depthgen}
     \mathbf{f}^p = Proj(E_p(\mathbf{p})).
\end{equation}

\begin{table*}[htb]
\centering
\resizebox{\linewidth}{!}{
\LARGE
\begin{tabular}{cccccccccccccccc}
\toprule
\multicolumn{1}{c}{\multirow{3}{*}{Method}} &\multicolumn{9}{c}{Scannet} &\multicolumn{6}{c}{S3DIS}\\
\cline{2-16}
    \multicolumn{1}{c}{} & \multicolumn{3}{c}{B15/N4} & \multicolumn{3}{c}{B12/N7} & \multicolumn{3}{c}{B10/N9} & \multicolumn{3}{c}{B8/N4} & \multicolumn{3}{c}{B6/N6} \\
\cline{2-10}  \cline{11-16} 
\multicolumn{1}{c}{} &hIoU & mIoU$^\mathcal{B}$ & mIoU$^\mathcal{N}$ &hIoU & mIoU$^\mathcal{B}$ &mIoU$^\mathcal{N}$ & hIoU &mIoU$^\mathcal{B}$ &mIoU$^\mathcal{N}$ & hIoU &mIoU$^\mathcal{B}$ &mIoU$^\mathcal{N}$ &hIoU & mIoU$^\mathcal{B}$ &mIoU$^\mathcal{N}$\\

\toprule
LSeg-3D~\cite{lis2024leg} &0.0 &64.4 &0.0 &0.9 &55.7 &0.1 &1.8 &68.4 &0.9 &0.1 &49.0 &0.1 &0.0 &30.1 &0.0\\
3DGenZ~\cite{michele2021generative} &20.6 &56.0 &12.6 &19.8 &35.5 &13.3 &12.0 &63.6 &6.6 &8.8 &50.3 &4.8 &9.4 &20.3 &6.1\\
3DTZSL~\cite{cheraghian2020transductive}&10.5 &36.7 &6.1 &3.8 &36.6 &2.0 &7.8 &55.5 &4.2 &8.4 &43.1 &4.7 &3.5 &28.2 &1.9\\
PLA~\cite{ding2023pla}&65.3 &68.3 &62.4 &55.3 &69.5 &45.9 &53.1 &76.2 &40.8 &34.6 &59.0 &24.5 &38.5 &55.5 &29.4\\
OpenScene~\cite{peng2023openscene}$^\dag$ &67.1 &68.8 &62.8 &56.8 &61.5 &51.7 &55.7 &71.8 &43.6 &39.1 &58.6 &33.2 &41.2 &56.2 &36.4\\
RegionPLC ~\cite{yang2023regionplc}&69.4 &68.2 &70.7 &68.2 &69.9 &66.6 &64.3 &76.3 &55.6 &- &- &- &- &- &-\\
OpenIns3D~\cite{huang2023openins3d}$^\dag$&72.6 &71.8 &73.4 &69.3 &70.7 &68.8 &64.5 &80.2 &49.8 &46.8 &63.2 &38.6 &53.6 &60.8 &47.5\\
\toprule
UniM-OV3D (Ours)&\textbf{75.8} &\textbf{75.3} &\textbf{76.6} &\textbf{74.7} &\textbf{75.2} &\textbf{74.1} &\textbf{69.9} &\textbf{83.5} &\textbf{57.3} &\textbf{54.6} &\textbf{68.3} &\textbf{45.2} &\textbf{59.1} &\textbf{66.7} &\textbf{54.2} \\ 
\toprule
\end{tabular}
}
\caption{Results for open-vocabulary 3D semantic segmentation on ScanNet and S3DIS. The evaluation metrics are hIoU, mIoU$^\mathcal{B}$ and mIoU$^\mathcal{N}$. $\dag$ denotes results reproduced by us on its official implementation. Best open-vocabulary results are highlighted in \textbf{bold}.}
\label{tab:1}
\end{table*}
\paragraph{Dense Associations across Modalities.}As shown in Figure~\ref{fig:2}, with a 3D scene and its corresponding features $\mathbf{f}^p$, $\mathbf{f}^t$, $\mathbf{f}^{img}$, $\mathbf{f}^d$, we employ a contrastive loss between point clouds and other modalities, which effectively compels the 3D embeddings to align with the joint representation space, formulated as:
\begin{align}
     {\mathcal{L}}_{(M_1,M_2)} =  \sum_{i,j}-\frac{1}{2}log{\frac{\exp(\mathbf{f}_i^{M_1} \cdot \mathbf{f}_j^{M_2} /\epsilon )}{\sum _{k}\exp(\mathbf{f}_i^{M_1} \cdot \mathbf{f}_k^{M_2}/\epsilon) } }  \nonumber\\
     - \frac{1}{2}log\frac{\exp(\mathbf{f}_i^{M_1} \cdot \mathbf{f}_j^{M_2} /\epsilon )}{\sum _{k}\exp(\mathbf{f}_k^{M_1} \cdot \mathbf{f}_j^{M_2}/\epsilon) } 
\end{align}
where $M_1$ and $M_2$ represent two modalities in $(point,image,depth)$ and $(i, j)$ indicates a positive pair in each training batch. And $ \epsilon$ is a learnable temperature parameter, similar to CLIP~\cite{radford2021learning}.

Finally, the overall objective function to perform unified multimodal alignment is defined by:
\begin{equation}\label{eq:capt_total}
     \mathcal{L}_{overall} = {\mathcal{L}}_{(P,I)}+ {\mathcal{L}}_{(P,D)} + {\mathcal{L}}_{(D,I)} + \mathcal{L}_{capt}^{total}
\end{equation}
where the language modality delivers comprehensive and scalable textual descriptions, while the image modality provides accurate guidance on object edges and contextual data. Moreover, the depth and 3D modality exposes vital structural details of objects. By aligning these modalities jointly in a common space, our approach can maximize the synergistic advantages among them, resulting in superior open-vocabulary scene understanding performance.

\section{Experiments}
\subsection{Setup}\label{sec:4.1}
\paragraph{Datasets and Metrics.}
To validate the effectiveness of our proposed UniM-OV3D, we conduct extensive experiments on four popular public 3D benchmarks: ScanNet~\cite{dai2017scannet}, ScanNet200~\cite{rozenberszki2022language}, S3DIS~\cite{armeni20163d} and nuScenes~\cite{caesar2020nuscenes}. The first three provide RGBD images and 3D meshes of indoor scenes, while the last one supplies Lidar scans of outdoor scenes. We use all four datasets to compare to alternative methods, covering both semantic and instance segmentation. For semantic segmentation, we employ the commonly used 3D segmentation metric mean intersection over union (mIoU$^\mathcal{B}$, mIoU$^\mathcal{N}$) and harmonic mean IoU (hIoU) for evaluating base, novel categories and their harmonic mean separately. For instance segmentation, We apply mean average precision under 50$\%$ IoU threshold (mAP$_{50} ^\mathcal{B}$, mAP$_{50} ^\mathcal{N}$, hAP$_{50}$) as major indicators.

\begin{table}[t]
\centering
\resizebox{\linewidth}{!}{
\LARGE
\begin{tabular}{ccccccc}
\toprule
\multicolumn{1}{c}{\multirow{3}{*}{Method}} &\multicolumn{6}{c}{ScanNet200}\\
\cline{2-7}
    \multicolumn{1}{c}{}    & \multicolumn{3}{c}{B170/N30} & \multicolumn{3}{c}{B150/N50} \\
\cline{2-7} 
\multicolumn{1}{c}{}  & hIoU &mIoU$^\mathcal{B}$ &mIoU$^\mathcal{N}$ &hIoU & mIoU$^\mathcal{B}$ &mIoU$^\mathcal{N}$\\
\toprule
3DGenZ~\cite{michele2021generative} &2.6 &15.8 &1.4 &3.3 &14.1 &1.9\\
3DTZSL~\cite{cheraghian2020transductive}&0.9 &4.0 &0.5 &0.7 &3.8 &0.4\\
LSeg-3D~\cite{lis2024leg}&1.5 &21.1 &0.8 &3.0 &20.6 &1.6 \\
PLA~\cite{ding2023pla} &11.4 &20.9 &7.8 &10.1 &20.9 &6.6\\
OpenScene~\cite{peng2023openscene}$^\dag$ &15.3 &22.5 &10.4 &16.8 &23.5 &11.2\\
RegionPLC~\cite{yang2023regionplc} &16.6 &21.6 &13.9 &14.6 &22.4 &10.8\\
OpenIns3D~\cite{huang2023openins3d}$^\dag$&17.2 &24.6 &13.6 &18.9 &25.8 &16.4 \\
\toprule
UniM-OV3D (Ours)&\textbf{22.3} &\textbf{29.5} &\textbf{17.1} &\textbf{25.8} &\textbf{30.7} &\textbf{21.6} \\
\toprule
\end{tabular}}
\caption{Results for open-vocabulary 3D semantic segmentation on ScanNet200 on hIoU, mIoU$^\mathcal{B}$ and mIoU$^\mathcal{N}$. $\dag$ denotes results reproduced by us on its official implementation.}
\label{tab:2}
\end{table}
\paragraph{Category Partition.}
ScanNet densely annotates 20 classes, ScanNet200 owns 200 classes, S3DIS contains 13 classes and nuScenes involves 16 classes. Following~\cite{ding2023pla,yang2023regionplc}, we disregard the “otherfurniture” class in ScanNet and randomly partition the rest 19 classes into 3 base/novel partitions, i.e. B15/N4 (15 base and 4 novel categories), B12/N7 and B10/N9, for semantic segmentation. Following SoftGroup~\cite{vu2022softgroup} to exclude two background classes, we acquire B13/N4, B10/N7, and B8/N9 partitions for instance segmentation on ScanNet. Similarly, we ignore the “clutter” class in S3DIS and get B8/N4, B6/N6 for both semantic and instance segmentation. For ScanNet200, we split 200 classes to B170/N30 and B150/N50. For nuScenes, we drop the “otherflat” class and obtain B12/N3 and B10/N5. 


\subsection{Main Results}\label{sec:4.2}
To validate the effectiveness of our proposed UniM-OV3D, we compare it with previous state-of-the-art 3D open-vocabulary methods on semantic and instance segmentation tasks.
\paragraph{3D Semantic Segmentation.} As shown in Table~\ref{tab:1}, compared to previous captioning-learning based method RegionPLC ~\cite{yang2023regionplc}, we obtain 5.1$\%$-13$\%$ mIoU performance gains among different partitions on ScanNet. Compared with previous zero-shot state-of-the-art method OpenIns3D~\cite{huang2023openins3d}, our method still obtains 3.2$\%$-5.4$\%$ and 5.5$\%$-7.8$\%$ improvements in terms of hIoU across various datasets and partitions. It is noteworthy that our approach consistently improves performance across different datasets and partitions. Moreover, for partitions with more novel categories, our method shows even greater enhancement. This further demonstrates the robustness and effectiveness of our approach in open vocabulary scene understanding tasks.

When facing the long-tail problems in ScanNet200, as shown in Table~\ref{tab:2}, our method still surpasses corresponding zero-shot method by 5.1$\%$-6.9$\%$ hIoU on both splits and 3.5$\%$-5.2$\%$ mIoU on novel categories. Furthermore, when evaluating the performance of our UniM-OV3D on outdoor setting, our method lifts the hIoU and mIoU of novel categories by 4.8$\%$-6.4$\%$ and 5$\%$-5.6$\%$ as shown in Table~\ref{tab:3}. In this regard, UniM-OV3D demonstrates superior generalizability to complex 3D open-world scenarios.
\begin{table}[t]
\centering
\resizebox{\linewidth}{!}{
\huge
\begin{tabular}{ccccccc}
\toprule
\multicolumn{1}{c}{\multirow{3}{*}{Method}} &\multicolumn{6}{c}{nuScenes}\\
\cline{2-7}
    \multicolumn{1}{c}{}    & \multicolumn{3}{c}{B12/N3} & \multicolumn{3}{c}{B10/N5} \\
\cline{2-7} 
\multicolumn{1}{c}{}  & hIoU &mIoU$^\mathcal{B}$ &mIoU$^\mathcal{N}$ &hIoU & mIoU$^\mathcal{B}$ &mIoU$^\mathcal{N}$\\
\toprule
3DGenZ~\cite{michele2021generative} &1.6 &53.3 &0.8 &1.9 &44.6 &1.0\\
3DTZSL~\cite{cheraghian2020transductive} &1.2 &21.0 &0.6 &6.4 &17.1 &3.9\\
LSeg-3D~\cite{lis2024leg} &0.6 &74.4 &0.3 &0.0 &71.5 &0.0 \\
PLA~\cite{ding2023pla} &47.7 &73.4 &35.4 &24.3 &73.1 &14.5\\
OpenScene~\cite{peng2023openscene}$^\dag$ &55.6 &75.4 &39.6 &28.5 &75.8 &19.2\\
RegionPLC~\cite{yang2023regionplc} &64.4 &75.8 &56.0 &49.0 &75.8 &36.3\\
OpenIns3D~\cite{huang2023openins3d}$^\dag$&65.4 &77.3 &59.6 &49.7 &78.7 &39.2 \\
\toprule
UniM-OV3D (Ours)&\textbf{70.2} &\textbf{80.5} &\textbf{64.6} &\textbf{55.1} &\textbf{81.7} &\textbf{44.8} \\
\toprule
\end{tabular}}
\caption{Results for open-vocabulary 3D semantic segmentation on nuScenes in terms of hIoU, mIoU$^\mathcal{B}$ and mIoU$^\mathcal{N}$. $\dag$ denotes results reproduced by us on its official implementation.}
\label{tab:3}
\end{table}

\begin{figure}
    \centering
\includegraphics[width=\linewidth]{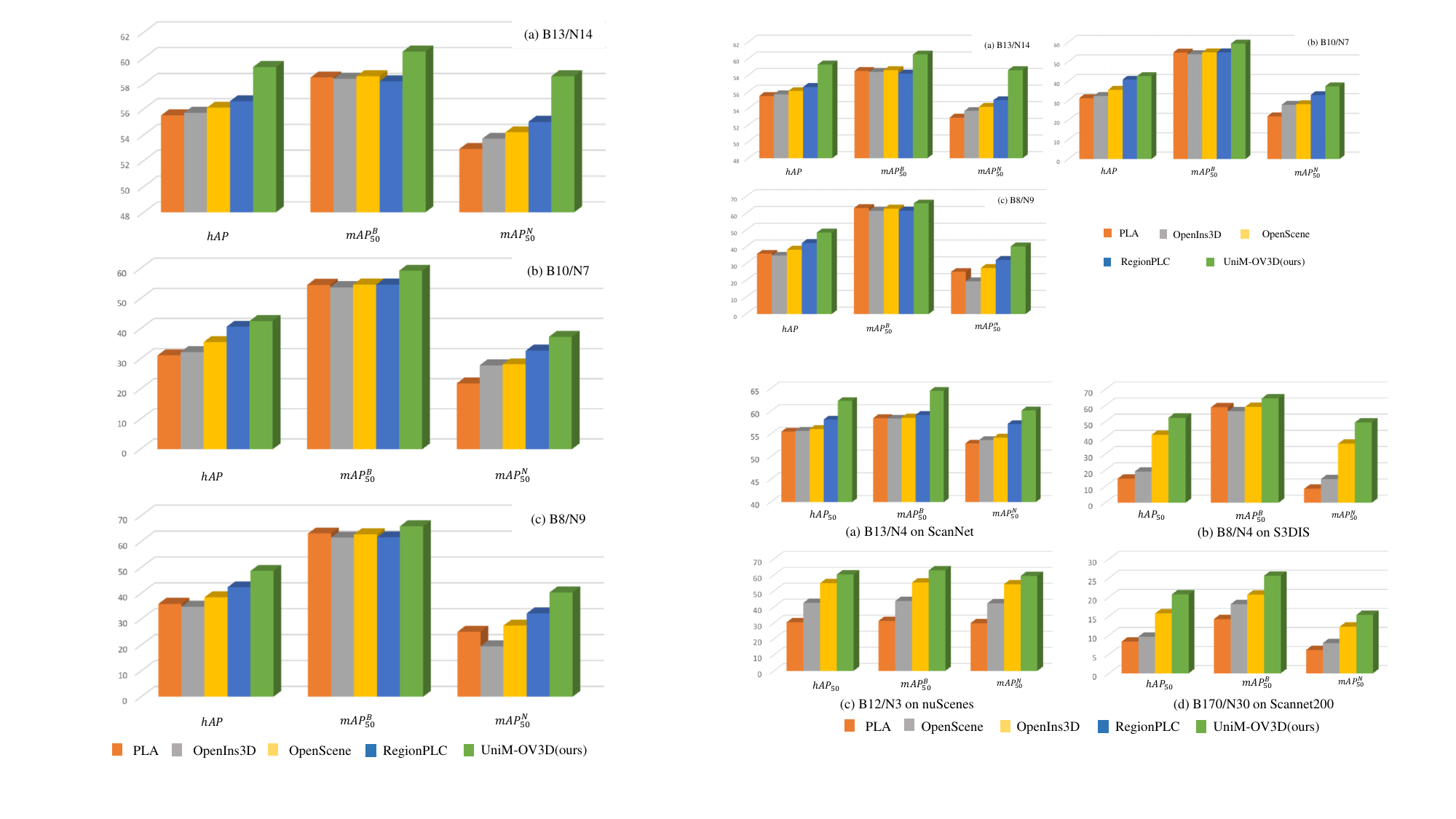}
 \vspace{-0.5cm}
    \caption{Comparisons on open-vocabulary 3D instance segmentation. (a) B13/N4 on ScanNet, (b) B8/N4 on S3DIS, (c) B12/N3 on nuScenes and (d) B170/N30 on ScanNet200.}
    \label{fig:5}
 \vspace{-0.4cm}
\end{figure}
\paragraph{3D Instance Segmentation.}
Our pipeline not only offers point-semantic language descriptions for fine-grained point clouds, but also encourages points to learn discriminative features, which in turn benefits the instance-level localization task. As shown in Figure~\ref{fig:5}, by learning a uni-modalities feature representations, our method outperforms RegionPLC~\cite{yang2023regionplc} by 4.1$\%$ hAP$_{50}$, 5.3$\%$ mAP$_{50}^\mathcal{B}$, 3$\%$ mAP$_{50}^\mathcal{N}$ on B13/N4 on the ScanNet. 
On the other datasets, our UniM-OV3D surpasses OpenIns3D~\cite{huang2023openins3d} by 5$\%$-10.6$\%$ hAP$_{50}$, 5$\%$-5.3$\%$ mAP$_{50}^\mathcal{B}$ and 3.1$\%$-13.2$\%$ mAP$_{50}^\mathcal{N}$ on their respective B8/N4, B12/N3 and B170/N30 partitions (RegionPLC only reports their instance segmentation results on ScanNet without open-source code). This illustrates the efficacy of our UniM-OV3D in empowering the model to identify unseen instances without the need for human annotations. Due to the space limition, specific indicators and more instance segmentation results are presented in the supplementary.

\subsection{Ablation Studies}\label{sec:4.3}
\begin{table}
\centering
\resizebox{\linewidth}{!}{
\begin{tabular}{ccccccccc}
\toprule
\multicolumn{1}{c}{\multirow{2}{*}{Uni}} & \multicolumn{1}{c}{\multirow{2}{*}{HFE}} &\multicolumn{1}{c}{\multirow{2}{*}{PCL}} &\multicolumn{3}{c}{B15/N4} & \multicolumn{3}{c}{B12/N7}\\
\cline{4-9} 
\multicolumn{1}{c}{}  & \multicolumn{1}{c}{}  & \multicolumn{1}{c}{}  & hIoU &mIoU$^\mathcal{B}$ &mIoU$^\mathcal{N}$ &hIoU & mIoU$^\mathcal{B}$ &mIoU$^\mathcal{N}$\\
\toprule
-  & -  & -  & 46.2 &57.1 &42.3 &43.5 &58.6 &41.0\\
- & -  & \ding{51}  & 62.1 &66.2 &56.4 &59.6 &67.9 &55.3\\
- & \ding{51}  & -  &60.0 &65.8 &56.2 &59.2 &66.5 &55.1 \\
\ding{51} & -  & -  & 60.3 &65.4 &57.6 &58.7 &66.2 &56.5\\
\toprule
\ding{51} & \ding{51}  & -  & 72.9 &72.8 &75.3 &72.6 &73.5 &71.3\\
\ding{51} & -  & \ding{51}  & 72.6 &72.1 &75.1 &71.9 &73.6 &71.8\\
- & \ding{51}  & \ding{51}  &74.8 &74.5 &76.0 &73.9 &74.6 &73.1 \\
\ding{51} & \ding{51}  & \ding{51}  & 75.8 &75.3 &76.6 &74.7 &75.2 &74.1\\
\toprule
\end{tabular}}
\caption{Ablations of different components of our network. Uni denotes the uni-modalities, HFE is the hierarchical point cloud feature extractor and PCL represents the point-semantic caption learning.}
\label{tab:4}
\end{table}
In this section, we change components and variants of our proposed UniM-OV3D by conducting extensive ablation studies on ScanNet dataset for semantic segmentation by default. And the partitions are mainly B15/N4 and B12/N7.
\begin{table}
\centering
\resizebox{\linewidth}{!}{
\begin{tabular}{ccccccc}
\toprule
 \multicolumn{1}{c}{\multirow{2}{*}{Fusion Method}} &\multicolumn{3}{c}{B15/N4} & \multicolumn{3}{c}{B12/N7}\\
\cline{2-7} 
  \multicolumn{1}{c}{}  & hIoU &mIoU$^\mathcal{B}$ &mIoU$^\mathcal{N}$ &hIoU & mIoU$^\mathcal{B}$ &mIoU$^\mathcal{N}$\\
\toprule
 Local PB  & 72.8	&72.4 &73.0 &71.6 &71.8 &71.3 \\
 Global PM  & 73.1 &72.6 &73.2 &71.7 &72.0 &71.4\\
 Local PB + Global PM  &75.0	&74.8 &75.3 &73.7 &73.8 &73.2 \\
 Local PB + Attn & 74.8	&74.6 &75.0 &73.6 &73.8 &73.1\\
 Global PM+ Attn  & 74.9 &74.8 &75.2 &73.8 &74.3 &73.5\\
 Local PB + Global PM + Attn  & 75.8 &75.3 &76.6 &74.7 &75.2 &74.1\\
\toprule
\end{tabular}}
\caption{Ablations of different fusion methods. Local PB denotes local PointBERT, Global PM is the global PointMAE and Attn represents the attention block.}
\label{tab:5}
\end{table}
\paragraph{Analysis of Different Components.} We illustrate the importance of different components of our network by removing some parts and keeping all the others unchanged. The baseline setting is to only use three modalities of image, text and point cloud. The 3D point cloud encoder uses sparse convolutional UNet~\cite{choy20194d}. The point captions come from corresponding image captions, and we use CLIP as the image and text encoder. As shown in Table~\ref{tab:4}, using the depth map to construct the uni-modality learning contributes much to the whole performance which proves the necessity of establishing a unified modal architecture. Without the point-semantic captions from different view, the performance drops dramatically which shows that coarse-to-fine supervision signal from fine-grained point-semantic caption learning is indeed a crucial factor. And the performance degradation caused by the absence of  hierarchical point cloud extractor proves that only the combination of local and global features can produce the best results.
\paragraph{Comparison of Different Fusion Methods in Hierarchical Feature Extractor.} To validate the effect of different fusion methods in the hierarchical point cloud feature extractor, we design six fusion patterns as shown in Table~\ref{tab:5}. With the attention block to transfer features between connected layers, the fused models override the original ones by 0.8$\%$-2.3$\%$ IoU. And our final fusion method of global and local with attention block outperforms the alternative by 0.5$\%$-1.4$\%$ mIoU.
\paragraph{Effect of Different Point-semantic Caption Views.}
\begin{table}
\centering
\resizebox{\linewidth}{!}{
\LARGE
\begin{tabular}{cccccccc}
\toprule
 \multicolumn{2}{c}{\multirow{2}{*}{Angle}} &\multicolumn{3}{c}{B15/N4} & \multicolumn{3}{c}{B12/N7}\\
\cline{3-8} 
  \multicolumn{2}{c}{}  & hIoU &mIoU$^\mathcal{B}$ &mIoU$^\mathcal{N}$ &hIoU & mIoU$^\mathcal{B}$ &mIoU$^\mathcal{N}$\\
\toprule
 \multicolumn{1}{c}{\multirow{4}{*}{sector-view}} &30  & 66.5	&68.4 &62.1 &67.7 &69.3 &61.3 \\
 \multicolumn{1}{c}{}  & 45  & 69.1 &69.4 &68.6 &68.3 &70.2 &66.3\\
 \multicolumn{1}{c}{}  &60  &72.3	&72.3 &74.2 &70.2 &73.3 &69.2 \\
 \multicolumn{1}{c}{}  &90  & 72.6	&72.3 &73.9 &70.1 &73.2 &69.1\\
 \toprule
 \multicolumn{1}{c}{\multirow{2}{*}{eye-view}} &120  & 73.1 &72.4 &74.3 &71.3 &73.4 &70.5\\
 \multicolumn{1}{c}{}  &180  & 71.9 &71.3 &73.6 &71.1 &73.1 &70.0\\
 \toprule
 \multicolumn{1}{c}{\multirow{1}{*}{global-view}} & 360  & 72.8 &72.7 &75.3 &72.6 &73.5 &71.3\\
\toprule
 \multicolumn{1}{c}{\multirow{3}{*}{combined-view}} & 360+60  & 73.3 &73.2 &75.6 &73.5 &74.2 &72.6\\
 & 360+120  & 74.2 &74.1 &75.7 &73.6 &74.3 &72.7\\
 & 360+60+120  &75.8 &75.3 &76.6 &74.7 &75.2 &74.1 \\
\toprule
\end{tabular}}
\caption{Performance of splitting point clouds into different angles and combining different views.}
\label{tab:6}
\end{table}
In order to explore the effect of dividing the point cloud into different angles in Section~\ref{sec:3.3} to obtain the corresponding fine-grained point-semantic description, we split the point cloud into six angles for experiments. As shown in Table~\ref{tab:6}, the 120 degree in {\tt eye-view} and 60 degree in {\tt sector-view} perform relatively well, but they are both slightly inferior to {\tt global-view} (360 degree). So, in order to further obtain the coarse-to-fine supervision, we combine 60 degree and 120 degree with 360 degree respectively, and finally adopt the method of combining the three in the last row.

\begin{table}
\setlength{\belowcaptionskip}{-0.5cm}
\centering
\resizebox{\linewidth}{!}{
\begin{tabular}{cccc}
\toprule
   Encoder  & CLIP & ImageBIND & PointBIND \\
\toprule
 hIoU/mIoU$^\mathcal{B}$/mIoU$^\mathcal{N}$  & 73.7/71.5/73.6&74.8/74.6/75.4 &75.8/75.3/76.6  \\
\toprule
\end{tabular}}
\caption{Performance of different image and text encoder on B15/N4 on ScanNet. }
\label{tab:7}
\end{table}
\paragraph{Text Encoder Selection.}
To extract the image and text embeddings, we experiment with different vision-language pre-trained encoders, CLIP~\cite{radford2021learning}, ImageBind~\cite{girdhar2023imagebind} and PointBIND~\cite{guo2023point}. As shown in Table~\ref{tab:7}, PointBIND exceeds the others which demonstrates that pre-training on point modality can provide better embeddings for 3D scene understanding.

\section{Conclusion}
In this paper, we propose UniM-OV3D, a unified muitimodal network for open-vocabulary 3D scene understanding. We jointly establish dense alignment between four modalities, \textit{i.e.}, point clouds, images, depth and text, to leverage their respective advantages. To fully capture local and global features of point clouds,  we stack spatial-aware layers to construct a hierarchical point cloud extractor. Furthermore, we deeply explore the possibility of directly using point clouds to generate corresponding captions, and the established point-semantic learning mechanism can provide language supervision signals more effectively. 
Extensive experiments on both indoor and outdoor datasets demonstrate the superiority of our model in 3D open-vocabulary scene understanding task. 

\bibliographystyle{named}
\bibliography{ijcai24}

\renewcommand\thefigure{A\arabic{figure}}
\renewcommand\thetable{A\arabic{table}}  
\renewcommand\theequation{A\arabic{equation}}
\setcounter{equation}{0}
\setcounter{table}{0}
\setcounter{figure}{0}
\appendix

\section{Overview}
This supplementary material includes:
\begin{itemize}
    \item Implementation details.
    \item Details about the attention block.
    \item Details of instance segmentation results.
    \item Qualitative results comparison. 
\end{itemize}

\section{Implementation Details}
We adopt PointLLM~\cite{xu2023pointllm} to generate point-semantic captions, the basic version of Vision Transformer~\cite{dosovitskiy2020image} with a patch size of 32 (namely ViT-B/32) as our depth encoder and SoftGroup~\cite{vu2022softgroup} for instance segmentation head. The intersecting angle between adjacent sectors is set to 0. During training, we use the AdamW~\cite{loshchilov2017decoupled} as the optimizer with an initial learning rate of 1$e-$4 and train for 180 epochs in semantic segmentation and 160 epochs in instance segmentation. We run all experiments with the batch size of 32 on 8 NVIDIA V100.

\section{Attention Block}
The attention block strives not just to highlight significant characteristics, but also to augment diverse depictions of entities at particular sites. Our attention block is structured as a bottom-up and top-down architecture. The bottom-up process is designed to gather comprehensive information, while the top-down process merges this comprehensive information with the initial feature maps. We employ the residual unit from ~\cite{he2016identity} as our fundamental component in the attention block. The attention block comprises three parts. In the first part, max pooling and a residual unit are utilized to expand the receptive field. Upon achieving the lowest resolution, a symmetrical top-down structure is devised to infer each pixel, thereby obtaining dense features in the second part. Moreover, we incorporate skip connections between the bottom-up and top-down feature maps to seize features at varying scales. In the third part, a bilinear interpolation is introduced following a residual unit to up-sample the output. Ultimately, we apply the sigmoid function to standardize the output after two successive $1\times 1$ convolution layers to maintain dimensional equilibrium.

\section{Instance Segmentation Results}
To illustrate the effectiveness of our proposed UniM-OV3D, we conduct instance segmentation on ScanNet~\cite{dai2017scannet}, ScanNet200~\cite{rozenberszki2022language}, S3DIS~\cite{armeni20163d} and nuScenes~\cite{caesar2020nuscenes}. Here we present specific metrics and full results.

As shown in Table~\ref{tab:supply1}, our method outperforms RegionPLC~\cite{yang2023regionplc} by 3.8$\%$-4.1$\%$ hAP$_{50}$, 5.3$\%$ mAP$_{50}^\mathcal{B}$, 2.9$\%$-5.8$\%$ mAP$_{50}^\mathcal{N}$ on different partitions on the ScanNet. As for on S3DIS, our method exceeds OpenIns3D~\cite{huang2023openins3d} by 10.6$\%$-10.8$\%$ hAP$_{50}$, 5.3$\%$-8.1$\%$ mAP$_{50}^\mathcal{B}$, 7.1$\%$-13.2$\%$ mAP$_{50}^\mathcal{N}$. RegionPLC only reports their instance segmentation results on ScanNet without open-source code. For the long-tail problems in ScanNet200, as shown in Table~\ref{tab:supply2}, we obtain 5$\%$-5.4$\%$ hAP$_{50}$, 3.9$\%$-5$\%$ mAP$_{50}^\mathcal{B}$, 3.1$\%$-4.2$\%$ mAP$_{50}^\mathcal{N}$ gains among different partitions. Furthermore, for outdoor scene, our method still lifts  hAP$_{50}$, mAP$_{50}^\mathcal{B}$ and mAP$_{50}^\mathcal{N}$ by 5.5$\%$-6.2$\%$,7.1$\%$-7.6$\%$ and 5.2$\%$-5.5$\%$ as shown in Table~\ref{tab:supply3}. The consistent performance in indoor and outdoor scenes in this instance segmentation task also proves the effectiveness and scalability of our unified modality network in 3D open-vocabulary scene understanding.

\begin{table}[htb]
\centering
\resizebox{\linewidth}{!}{
\LARGE
\begin{tabular}{ccccccc}
\toprule
\multicolumn{1}{c}{\multirow{3}{*}{Method}} &\multicolumn{6}{c}{ScanNet200}\\
\cline{2-7}
    \multicolumn{1}{c}{}    & \multicolumn{3}{c}{B170/N30} & \multicolumn{3}{c}{B150/N50} \\
\cline{2-7} 
\multicolumn{1}{c}{}  &hAP$_{50}$ & mAP$_{50}^\mathcal{B}$ & mAP$_{50}^\mathcal{N}$ & hAP$_{50}$ & mAP$_{50}^\mathcal{B}$ &mAP$_{50}^\mathcal{N}$\\
\toprule
PLA~\cite{ding2023pla} &8.3 &14.2 &6.1 &7.6 &16.2 &5.8\\
OpenScene~\cite{peng2023openscene} &9.6 &18.1 &7.9 &8.1 &17.3 &6.6\\
OpenIns3D~\cite{huang2023openins3d}&15.7 &20.6 &12.2 &16.5 &22.8 &14.3 \\
\toprule
UniM-OV3D (Ours)&\textbf{20.7} &\textbf{25.6} &\textbf{15.3} &\textbf{21.9} &\textbf{26.7} &\textbf{18.5} \\
\toprule
\end{tabular}}
\caption{Results for open-vocabulary 3D instance segmentation on ScanNet200 on hAP$_{50}$, mAP$_{50}^\mathcal{B}$ and mAP$_{50}^\mathcal{N}$. }
\label{tab:supply2}
\end{table}

\begin{table}[htb]
\centering
\resizebox{\linewidth}{!}{
\huge
\begin{tabular}{ccccccc}
\toprule
\multicolumn{1}{c}{\multirow{3}{*}{Method}} &\multicolumn{6}{c}{nuScenes}\\
\cline{2-7}
    \multicolumn{1}{c}{}    & \multicolumn{3}{c}{B12/N3} & \multicolumn{3}{c}{B10/N5} \\
\cline{2-7} 
\multicolumn{1}{c}{}  & hAP$_{50}$ & mAP$_{50}^\mathcal{B}$ & mAP$_{50}^\mathcal{N}$ & hAP$_{50}$ & mAP$_{50}^\mathcal{B}$ &mAP$_{50}^\mathcal{N}$\\
\toprule
PLA~\cite{ding2023pla} &30.3 &31.2 &29.8 &31.5 &32.6 &31.0\\
OpenScene~\cite{peng2023openscene} &42.5 &43.6 &42.1 &43.5 &44.2 &42.9\\
OpenIns3D~\cite{huang2023openins3d} &54.7 &55.2 &54.1 &55.1 &56.7 &54.5 \\
\toprule
UniM-OV3D (Ours)&\textbf{60.2} &\textbf{62.8} &\textbf{59.3} &\textbf{61.3} &\textbf{63.8} &\textbf{60.0} \\
\toprule
\end{tabular}}
\caption{Results for open-vocabulary 3D instance segmentation on nuScenes in terms of hAP$_{50}$, mAP$_{50}^\mathcal{B}$ and mAP$_{50}^\mathcal{N}$.}
\label{tab:supply3}
\end{table}

\begin{figure*}[t]
\centering
\includegraphics[width=0.95\linewidth]{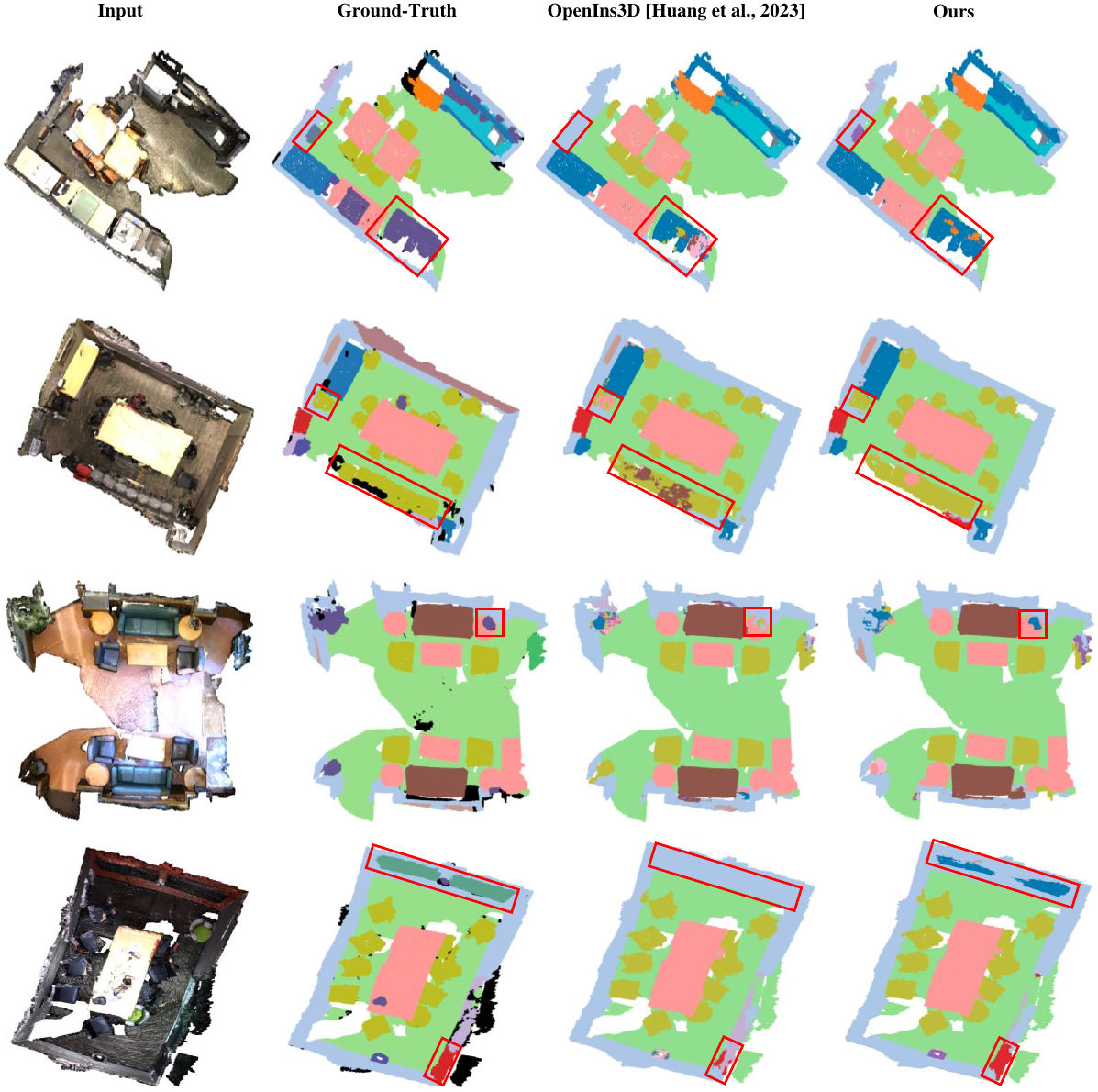}
\caption{Qualitative results comparison between our proposed UniM-OV3D and the state-of-the-art OpenIns3D~\protect\cite{huang2023openins3d} for semantic segmentation on ScanNet dataset.}
\label{fig:supply1}
\end{figure*}

\section{Quantitative Results}
The representative quantitative segmentation results of our proposed UniM-OV3D on the ScanNet dataset are shown in Figure~\ref{fig:supply1}. We also present the quantitative results of the state-of-the-art OpenIns3D~\protect\cite{huang2023openins3d}. We can observe that OpenIns3D exhibits poor performance in segmenting relative objects, incorrectly segments some objects in ground-truth, and even ignores objects that lacked spatial structure in ground-truth. Our UniM-OV3D makes full use of the information of each modality and performs better in these segmentation scenarios.
\begin{table*}[htb]
\centering
\resizebox{\linewidth}{!}{
\LARGE
\begin{tabular}{cccccccccccccccc}
\toprule
\multicolumn{1}{c}{\multirow{3}{*}{Method}} &\multicolumn{9}{c}{Scannet} &\multicolumn{6}{c}{S3DIS}\\
\cline{2-16}
    \multicolumn{1}{c}{} & \multicolumn{3}{c}{B13/N4} & \multicolumn{3}{c}{B10/N7} & \multicolumn{3}{c}{B8/N9} & \multicolumn{3}{c}{B8/N4} & \multicolumn{3}{c}{B6/N6} \\
\cline{2-10}  \cline{11-16} 
\multicolumn{1}{c}{} & hAP$_{50}$ & mAP$_{50}^\mathcal{B}$ & mAP$_{50}^\mathcal{N}$ & hAP$_{50}$ & mAP$_{50}^\mathcal{B}$ &mAP$_{50}^\mathcal{N}$ &  hAP$_{50}$ & mAP$_{50}^\mathcal{B}$ & mAP$_{50}^\mathcal{N}$ &  hAP$_{50}$ & mAP$_{50}^\mathcal{B}$ & mAP$_{50}^\mathcal{N}$ &  hAP$_{50}$ & mAP$_{50}^\mathcal{B}$ & mAP$_{50}^\mathcal{N}$\\

\toprule
LSeg-3D~\cite{lis2024leg} &5.1 &57.9 &2.6 &2.0 &50.7 &1.0 &2.4 &59.4 &1.2 &- &- &- &- &- &-\\
PLA~\cite{ding2023pla}&55.5 &58.5 &52.9 &31.2 &54.6 &21.9 &35.9 &63.1 &25.1 &15.0 &59.0 &8.6 &16.0 &46.9 & 9.8\\
OpenIns3D~\cite{huang2023openins3d}&55.7 &58.4 &53.7 &32.3 &53.8 &27.9 &34.8 &61.5 &19.5 &42.0 &59.2 &36.5 &33.0 &46.1 &31.3\\
OpenScene~\cite{peng2023openscene} &56.1 &58.6 &54.2 &35.6 &54.7 &28.2 &38.4 &62.8 &27.5 &19.3 &56.5 &14.7 &16.1 &47.2 &15.4\\
RegionPLC ~\cite{yang2023regionplc}&58.2 &59.2 &57.2 &40.6 &53.9 &32.5 &46.8 &62.5 &37.4 &- &- &- &- &- &-\\

\toprule
UniM-OV3D (Ours)&\textbf{62.3} &\textbf{64.5} &\textbf{60.2} &\textbf{44.6} &\textbf{59.3} &\textbf{38.3} &\textbf{50.6} &\textbf{67.9} &\textbf{40.3} &\textbf{52.6} &\textbf{64.5} &\textbf{49.7} &\textbf{43.8} &\textbf{54.2} &\textbf{38.4} \\ 
\toprule
\end{tabular}
}
\caption{Results for open-vocabulary 3D instance segmentation on ScanNet and S3DIS. The evaluation metrics are hAP$_{50}$, mAP$_{50}^\mathcal{B}$ and mAP$_{50}^\mathcal{N}$. Best open-vocabulary results are highlighted in \textbf{bold}.}
\label{tab:supply1}
\end{table*}

\end{document}